\documentclass[a4paper]{article}

\usepackage{INTERSPEECH2019}
\usepackage{url}
\DeclareMathOperator{\Att}{Att}

\DeclareMathOperator{\softmax}{softmax}

\title{Language learning using Speech to Image retrieval}
\name{Danny Merkx, Stefan L. Frank, Mirjam Ernestus}
\address{
  Centre for Language Studies, Radboud University, Nijmegen, The Netherlands}
\email{d.merkx@let.ru.nl, s.frank@let.ru.nl, m.ernestus@let.ru.nl}

\begin{document}

\maketitle
\begin{abstract}

Humans learn language by interaction with their environment and listening to other humans. It should also be possible for computational models to learn language directly from speech but so far most approaches require text. We improve on existing neural network approaches to create visually grounded embeddings for spoken utterances. Using a combination of a multi-layer GRU, importance sampling, cyclic learning rates, ensembling and vectorial self-attention our results show a remarkable increase in image-caption retrieval performance over previous work. Furthermore, we investigate which layers in the model learn to recognise words in the input. We find that deeper network layers are better at encoding word presence, although the final layer has slightly lower performance. This shows that our visually grounded sentence encoder learns to recognise words from the input even though it is not explicitly trained for word recognition. 

\end{abstract}


\section{Introduction}

Most computational models of natural language processing (NLP) are based on written language; machine translation, sentence meaning representation and language modelling to name a few (e.g. \cite{neubig2018, kiros2015}). Even if the task inherently involves speech, such as in automatic speech recognition, models require large amounts of transcribed speech \cite{wang2016}. Yet, humans are capable of learning language from raw sensory input, and furthermore children learn to communicate long before they are able to read. In fact, many languages have no orthography at all and there are also languages of which the writing system is not widely used by its speakers. Text-based models cannot be used for these languages and applications like search engines and automated translators cannot serve these populations. 

There has been increasing interest in learning language from more natural input, such as directly from the speech signal, or multi-modal input (e.g. speech and vision). This has several advantages such as removing the need for expensive annotation of speech, being applicable to low resource languages and being more plausible as a model of human language learning.

An important challenge in learning language from spoken input is the fact that the input is not presented in neatly segmented tokens. An auditory signal does not contain neat breaks in between words like the spaces in text. Furthermore, no two realisations of the same spoken word are ever exactly the same. As such, spoken input cannot be represented by conventional word embeddings (e.g. word2vec \cite{Mikolov2013}, GloVe \cite{Pennington2014}). These text-based embeddings are trained to encode word-level semantic knowledge and have become a mainstay in work on sentence representations (e.g. \cite{Conneau2017, Kiela2018}). When we want to learn language directly from speech, we will have to do so in a more end-to-end fashion, without prior lexical level knowledge in terms of both form and semantics. 

In previous work \cite{merkx2019} we used image-caption retrieval, where given a written caption the model must return the matching image and vice versa. We trained deep neural networks (DNNs) to create sentence embeddings without the use of prior knowledge of lexical semantics  (see \cite{Kiela2018, Karpathy2017, Faghri2018} for other studies on this task). The visually grounded sentence embeddings that arose capture semantic information about the sentence as measured by the Semantic Textual Similarity task (see \cite{Agirre2016}), performing comparably to text-only methods that require word embeddings. 

In the current study we present an image-caption retrieval model that extends our previous work to spoken input. In \cite{Harwath2015, Harwath2016}, the authors adapted text based caption-image retrieval (e.g. \cite{Karpathy2017}) and showed that it is possible to perform speech-image retrieval using convolutional neural networks on spectral features. Our work is most closely related to the models presented in \cite{Harwath2015, Harwath2016, Harwath2018, Chrupala2017}. In the current study we improve upon these previous approaches to visual grounding of speech and present state-of-the-art image-caption retrieval results.

The work by \cite{Harwath2015, Harwath2016, Harwath2018, Chrupala2017} and the results presented here are a step towards more cognitively plausible models of language learning as it is more natural to learn language without prior assumptions about the lexical level. For instance, research indicates that the adult lexicon contains many relatively fixed multi-word expressions (e.g., `how-are-you-doing') \cite{Tomasello}. Furthermore, early during language acquisition the lexicon consists of entire utterances before a child's language use becomes more adult-like \cite{Tomasello, Braine, Pine1993, lieven2003}. Image to spoken-caption retrieval models do not know a priori which constituents of the input are important and have no prior knowledge of lexical level semantics. We probe the resulting model to investigate whether it learns to recognise lexical units in the input without being explicitly trained to do so. 

We test two types of acoustic features; Mel Frequency Cepstral Coefficients (MFCCs) and Multilingual Bottleneck (MBN) features. MFCCs are features that can be computed for any speech signal without needing any other data, while the MBN features are `learned' features that result from training a network on top of MFCCs in order to recognise phoneme states. While MBN features have been shown to be useful in several speech recognition tasks (e.g. \cite{quoc2014, Fer:CSL:2017}), learned audio features face the same issue as word embeddings, as humans learn to extract useful features from the audio signal as a result of learning to understand language and not as a separate process. However, the MBN features can still be useful where system performance is more important than cognitive plausibility, for instance in a low resource setting. Furthermore, these features could provide a clue as to what performance would be possible if we had more sophisticated models or more data to improve the feature extraction from the MFCCs in an end-to-end fashion. 

In summary, we improve on previous spoken-caption to image retrieval models and investigate whether it learns to recognise words in the speech signal. We show that our model achieves state-of-the-art results on the Flickr8k database, outperforming previous models by a large margin using both MFCCs and MBN features. We find that our model learns to recognise words in the input signal and show that the deeper layers are better at encoding this information. Recognition performance drops a little in the last two layers as the network abstracts away from the detection of specific words in the input and learns to map the utterances to the joint embedding space. We released the code for this project on github:  \url{https://github.com/DannyMerkx/speech2image/tree/Interspeech19}.

\section{Image to spoken-caption retrieval}

\subsection{Materials}

Our model is trained on the Flickr8k database \cite{Hodosh2015}. Flickr8k contains 8,000 images taken from online photo sharing application Flickr.com, for which five English captions per image are available. Annotators were  asked  to  ‘write  sentences  that  describe  the  depicted scenes,  situations,  events  and  entities  (people,  animals,  other  objects)’. Spoken captions for Flickr8k were collected by \cite{Harwath2015} by having Amazon Mechanical Turk workers pronounce the original written captions. We used the data split provided by \cite{Karpathy2017}, with 6,000 images for training and a development and test set both of 1,000 images.

\subsection{Image and acoustic features}

To extract image features, all images are resized such that the smallest side is 256 pixels while keeping the aspect ratio intact. We take ten 224 by 224 crops of the image: one from each corner, one from the middle and the same five crops for the mirrored image. We use ResNet-152 \cite{He2015} pretrained on ImageNet to extract visual features from these ten crops and then average the features of the ten crops into a single vector with 2,048 features.

We test two types of acoustic features; Mel Frequency Cepstral Coefficients (MFCCs) and Multilingual Bottleneck (MBN) features. The MFCCs were created using 40 Mel-spaced filterbanks. We use 12 MFCCs and the log energy feature and add the first and second derivatives resulting in 39-dimensional feature vectors. We compute the MFCCs using 25 ms analysis windows with a 5 ms shift.

The MBN features are created using a pre-trained DNN made available by \cite{Fer:CSL:2017}. In short, the network is trained on multilingual speech data (11 languages, no English) to classify phoneme states. The MBN features consist of the outputs of intermediate network layers where the network is compressed from 1500 features to 30 features (see \cite{Fer:CSL:2017} for the full details of the network and training). 

\subsection{Model architecture}

\begin{figure}
    \centering
    \includegraphics[width =\linewidth]{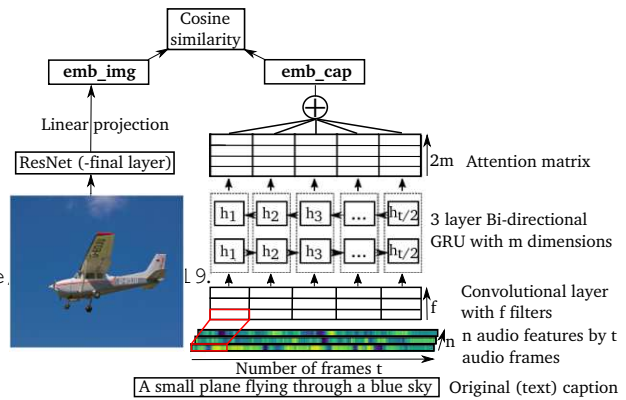}
    \caption{Model architecture: The model consists of two branches with the image encoder on the left and the caption encoder on the right. The audio features consist of $n$ features by $t$ frames and the RNN hidden states by $\mathbf{h}_{t/2}$. Each RNN hidden state has $m$ features which are concatenated for the forward and backward RNN into $2m$ dimensional hidden states. Vectorial attention is applied which weighs and sums the hidden states resulting in the caption embedding. At the top we calculate the cosine similarity between the image and caption embedding ($\textbf{emb\_img}$ and $\textbf{emb\_cap}$).}
    \label{network}
\end{figure}

Our multimodal encoder maps images and their corresponding captions to a common embedding space. The idea is to make matching images and captions lie close together and mismatched images and captions lie far apart in the embedding space. Our model consists of two parts; an image encoder and a sentence encoder as depicted in Figure \ref{network}. The approach is based on our own text-based model described in \cite{merkx2019} and on the speech-based models described in \cite{Harwath2016, Chrupala2017} and we refer to those studies for more details. Here, we focus on the differences with previous work.

For the image encoder we use a single-layer linear projection on top of the pretrained image recognition model, and normalise the result to have unit L2 norm. The image encoder has 2048 input units and 2048 output units.

Our caption encoder consists of three main components. First we apply a 1-dimensional convolutional layer to the acoustic input features. The convolution has a stride of size 2, kernel size 6 and 64 output channels. This is the only layer where the model differs from the text-based model, which features a character embedding layer instead of a convolutional layer. The resulting features are then fed into a bi-directional Gated Recurrent Unit (GRU) followed by a self-attention layer and is lastly normalised to have unit L2 norm. 

We use a 3-layer bi-directional GRU which allows the network to capture long-range dependencies in the acoustic signal (see \cite{Chung2014} for a more detailed description of the GRU). Furthermore, by making the layer bi-directional we let the network process the output of the convolutional layer from left to right and vice versa, allowing the model to capture dependencies in both directions. We use a GRU with 1024 units, and concatenate the bidirectional representations resulting in hidden states of size 2048. Finally, the self-attention layer computes a weighted sum over all the hidden GRU states: 
\begin{equation}
    \mathbf{a}_t = \softmax(V\tanh(W\mathbf{h}_t +\mathbf{b}_w)+\mathbf{b}_v) 
    \label{eq3}
\end{equation}
\begin{equation}
    \Att(\mathbf{h}_1, ..., \mathbf{h}_t) = \sum\limits_{t}\mathbf{a}_t\circ\mathbf{h}_t
    \label{eq4}
\end{equation}
where $\mathbf{a}_t$ is the attention vector for hidden state $\mathbf{h}_t$ and $W$, $V$, $\mathbf{b}_w$, and $\mathbf{b}_v$ indicate the weights and biases. The applied attention is then the sum over the Hadamard product between all hidden states $(\mathbf{h}_1, ..., \mathbf{h}_t)$ and their attention vector. We use 128 units for $W$ and 2048 units for $V$.

\subsection{Training}

Following \cite{merkx2019}, the model is trained to embed the images and captions such that the cosine similarity between image and caption pairs is larger (by a certain margin) than the similarity between mismatching pairs. This so called hinge loss $L$ as a function of the network parameters \(\theta\) is given by: 

\begin{equation}
    \begin{aligned}L(\theta) = \sum\limits_{(c,i),(c',i')\in B} \biggl(\max(0, \cos(c,i') - \cos(c,i) + \alpha) +\\
    \max(0, \cos(i,c') - \cos(i,c) + \alpha)\biggr) \end{aligned}
\end{equation}
where $(c,i)\neq(c',i')$. $B$ is a minibatch of correct caption-image pairs $(c,i)$, where the other caption-image pairs in the batch serve to create mismatched pairs $(c,i')$ and $(c',i)$. We take the cosine similarity $\cos(x,y)$ and subtract the similarity of the mismatched pairs from the matching pairs such that the loss is only zero when the matching pair is more similar than the mismatched pairs by a margin $\alpha$. We use importance sampling to select the mismatched pairs; rather than using all the other samples in the mini-batch as mismatched pairs (as done in \cite{merkx2019, Chrupala2017}), we calculate the loss using only the hardest examples (i.e. mismatched pairs with high cosine similarity). While \cite{Faghri2018} used only the single hardest example in the batch for text-captions, we found that this did not work for the spoken captions. Instead we found that using the hardest 25 percent worked well. 

The networks are trained using Adam \cite{Kingma2015} with a cyclic learning rate schedule based on \cite{Smith2015}. The learning rate schedule varies the learning rate smoothly between a minimum and maximum bound which were set to $10^{-6}$ and $2\times10^{-4}$ respectively. The learning rate schedule causes the network to visit several local minima during training, allowing us to use snapshot ensembling \cite{Huang2017}. By saving the network parameters at each local minimum, we can ensemble the embeddings of multiple networks at no extra cost. We use a margin $\alpha = 0.2$ for the loss function. We train the networks for 32 epochs and take a snapshot for ensembling at every fourth epoch. For ensembling we use the two snapshots with the highest performance on the development data and simply sum their embeddings. 

The main differences with the approaches described in \cite{Harwath2016, Chrupala2017} are the use of multi-layered GRUs, importance sampling, the cyclic learning rate, snapshot ensembling and the use of vectorial rather than scalar attention. 

\section{Word presence detection}

While our model is not explicitly trained to recognise words or segment the speech signal, previous work has shown that such information can be extracted by visual grounding models \cite{Chrupala2017, Kamper2017}. \cite{Chrupala2017} use a binary decision task: given a word and a sentence embedding, decide if the word occurs in the sentence. Our approach is similar to the spoken-bag-of-words prediction task described in \cite{Kamper2017}. Given a sentence embedding created by our model, a classifier has to decide which of the words in its vocabulary occur in the sentence. 

Based on the original written captions, our database contains 7,374 unique words with a combined occurrence frequency of 324,480. From these we select words that occur between 50 and a 1,000 times and are over 3 characters long so that there are enough examples in the data that the model might actually learn to recognise them, and to filter out punctuation, spelling mistakes, numerals and most function words. This leaves 460 unique words, mostly verbs and nouns, with a combined occurrence frequency of 87,020 in our data. We construct a vector for each sentence in Flickr8k indicating which of these words is present. We do not encode multiple occurrences of the same word in one sentence. 

The words described above are used as targets for a neural network classifier consisting of a single feed forward layer with 460 units. This layer simply takes an embedding vector as input and maps it to the 460 target words. We then apply the standard logistic function and calculate the Binary Cross Entropy loss to train the network. 

We train five word detection networks for both the MFCC and the MBN based caption encoders, in order to see how word presence is encoded in the different neural network layers. We train networks for the final output layer, the three intermediate layers of the GRU and the acoustic features. For the final layer we simply use the output embedding as input to the word detection network. We apply some post-processing to the acoustic features and the intermediate layer outputs to ensure that our word detection inputs are all of the same size. As the intermediate GRU layers produce 2048 features for each time step in the signal, we use average-pooling along the temporal dimension to create a single input vector and normalise the result to have unit L2 norm. The acoustic features consist of 30 (MBN) or 39 (MFCC) features for each time step, so we apply the convolutional layer followed by an untrained GRU layer to the input features, use average-pooling and normalise the result to have unit L2 norm. 

The word detection networks are trained for 32 epochs using Adam \cite{Kingma2015} with a constant learning rate of 0.001. We use the same data split that was used for training the multi-modal encoder, so that we test word presence detection on data that was not seen by either the encoder or the decoder. 

\section{Results}

\begin{table}
    \caption{Image-Caption retrieval results on the Flickr8k test set. R@N is the percentage of items for which the correct image or caption was retrieved in the top N (higher is better). Med r is the median rank of the correct image or caption (lower is better). We also report the 95 percent confidence interval for the R@N scores.}
    \centering
    \resizebox{\linewidth}{!}{
        \begin{tabular}{l | r r r r }
        \toprule
            \multicolumn{1}{l}{Model} & \multicolumn{4}{c}{Caption to Image} \\
            \midrule
             & R@1 & R@5 & R@10 & med r \\ 
            \cite{Harwath2015} & - & - & 17.9$\pm$1.1& - \\
            \cite{Chrupala2017} & 5.5$\pm$0.6 & 16.3$\pm$1.0 & 25.3$\pm$1.2 & 48 \\
            MFCC-GRU & 8.4$\pm$0.8 & 25.7$\pm$1.2 & 37.6$\pm$1.3 & 21 \\ 
            MBN-GRU & 12.7$\pm$0.9 & 34.9$\pm$1.3 & 48.5$\pm$1.4 & 11\\
            Char-GRU \cite{merkx2019}& 27.5$\pm$1.2 & 58.2$\pm$1.4 & 70.5$\pm$1.3 & 4 \\
            \toprule
            \multicolumn{1}{l}{Model} & \multicolumn{4}{c}{Image to Caption} \\
            \midrule
             & R@1 & R@5 & R@10 & med r \\
            \cite{Harwath2015} & - & - & 24.3$\pm$2.7 & - \\
            MFCC-GRU & 12.2$\pm$2.0 & 31.9$\pm$2.9 & 45.2$\pm$3.1 & 13 \\ 
            MBN-GRU & 16.0$\pm$2.5& 42.8$\pm$3.1& 56.1$\pm$3.0 & 8\\
            Char-GRU \cite{merkx2019} & 38.5$\pm$3.0 & 68.9$\pm$2.9 & 79.3$\pm$2.5 & 2\\
            \bottomrule
        \end{tabular}
    \label{flickr_c2i_results}
    }
\end{table}

\begin{figure}
    \centering
    \includegraphics[width =.9\linewidth]{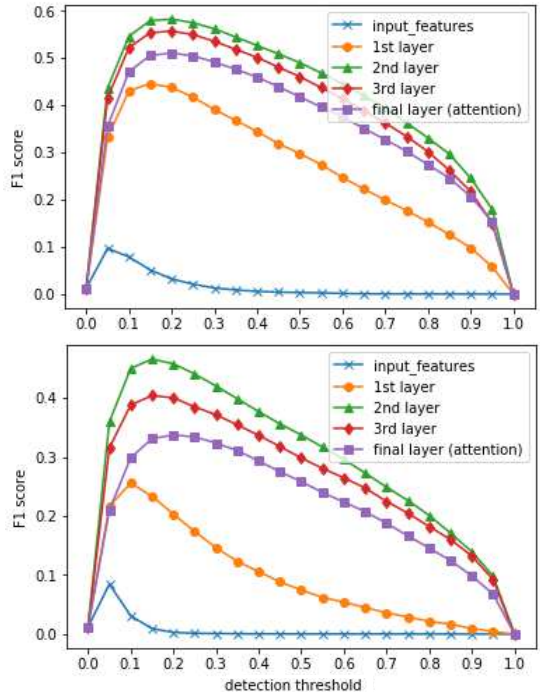}
    \caption{Plots of the F1 scores for the word presence classifiers at 20 equally spaced activation thresholds. The top figure shows the classifiers trained on the MBN model, and the bottom figure the MFCC model.}
    \label{probing}
\end{figure}

Table \ref{flickr_c2i_results} shows the performance of our models on the image-caption retrieval task. The caption embeddings are ranked by cosine distance to the image and vice versa where R@N is the percentage of test items for which the correct image or caption was in the top N results. We compare our models to \cite{Harwath2015} and \cite{Chrupala2017}, and include our own character-based model for comparison. \cite{Harwath2015} is a convolutional approach, whereas \cite{Chrupala2017} is an approach using recurrent highway networks with scalar attention. The character-based model is similar to the model we use here and was trained on the original Flickr8k text captions (see \cite{merkx2019} for a full description). Both our MFCC and MBN based model significantly outperform previous spoken caption-to-image methods on the Flickr8k dataset. The largest improvement is the MBN model which outperforms the results reported in \cite{Chrupala2017} by as much as 23.2 percentage points on R@10. The MFCC model also improves on previous results but scores significantly lower than the MBN model across the board, improving as much as 12.3 percentage points over previous work. There is a large performance gap between the text-caption to image retrieval results and the spoken-caption to image results, showing there is still a lot of room for improvement.

The results of the word presence detection task are shown in Figure \ref{probing} and Table \ref{auc}. Figure \ref{probing} shows the F1 score for all the classifiers at 20 equally spaced detection thresholds (i.e. a word is classified as `present' if the word detection output is above this threshold). Table~\ref{auc} displays the area under the curve for the receiver operating characteristic. Even though the MBN model outperforms the MFCC model for all layers we see the same pattern emerging from both the F1 score and the AUC. The performance on the feature level is not much better than random. Predicting `not present'  for every word would be the best random guess as this is a heavy majority class in this task. Inspection of the predictions shows that the classifier is indeed heavily biased towards the majority class for the input features. Then we see the performance increasing for the first layer and peaking at the second layer. The performance then drops slightly for the third layer and the attention layer. 

\begin{table}
    \caption{Area under the curve of the receiver operating characteristic for both models. }
    \centering
    \resizebox{\linewidth}{!}{
        \begin{tabular}{l | r r r r r}
        \toprule
            \multicolumn{1}{l}{Model} & \multicolumn{5}{c}{AUC} \\
            \midrule
             & input & layer 1 & layer 2 & layer 3 & attention \\ 
            MBN & .57 & .80 & .86 & .85 & .82 \\
            MFCC & .54 & .68 & .80 & .75 & .75 \\
            \bottomrule
        \end{tabular}
    \label{auc}
    }
\end{table}

\section{Discusion and Conclusion}

We trained an image-caption retrieval model on spoken input and investigated whether it learns to recognise linguistic units in the input. As improvements over previous work we used a 3-layer GRU and employed importance sampling, cyclic learning rates, ensembling and vectorial self-attention. Our results on both MBN and MFCC features are significantly higher than the previous state-of-the-art. 
The largest improvement comes from using the learned MBN features but our approach also improves results for MFCCs, which are the same features as were used in \cite{Chrupala2017}. The learned MBN features provide better performance whereas the MFCCs are more cognitively plausible input features.

The probing task shows that the model learns to recognise these words in the input. The system is not explicitly optimised to do so, but our results show that the lower layers learn to recognise this form related information from the input. After layer 2, the performance starts to decrease slightly which might indicate that these layers learn a more task-specific representation and it is to be expected that the final attention layer specialises in mapping from audio features to the multi-modal embedding space. 

In conclusion, we presented what are, to the best of our knowledge, the best results on spoken-caption to image retrieval. Our results improve significantly over previous approaches for both untrained and trained audio features. In a probing task, we show that the model learns to recognise words in the input speech signal. 

We are currently collecting the Semantic Textual Similarity (STS) database in spoken format and the next step will be to investigate whether the model presented here also learns to capture sentence level semantic information and understand language in a deeper sense than recognising word presence. The work presented in \cite{Chrupala2017} has made the first efforts in this regard and we aim to extend this to a larger database with sentences from multiple domains. Furthermore, we want to investigate the linguistic units that our model learns to recognise. In the current study, we only investigated whether the model learns to recognise words, but the potential benefit of our model is that it might learn multi-word statements or might even learn to look at sub-lexical level information. \cite{Harwath2018, Drexler2017} have recently shown that the speech-to-image retrieval approach can be used to detect word boundaries and even discover sub-word units. Our interest is in investigating how these word and sub-word units develop over training and through the network layers. 

\section{Acknowledgements}

The research presented here was funded by the Netherlands Organisation for Scientific Research (NWO) Gravitation Grant 024.001.006 to the Language in Interaction Consortium. This work was carried out on the Dutch national e-infrastructure with the support of SURF Cooperative.

\bibliographystyle{IEEEtran}

\bibliography{mybib}


\end{document}